\documentclass[runningheads]{llncs}

\usepackage{eccv}

\usepackage{eccvabbrv}
\usepackage{amssymb}
\usepackage{graphicx}
\usepackage{booktabs}
\usepackage{multirow}
\usepackage{makecell}
\usepackage{mathtools}
\usepackage{amssymb}

\usepackage{array}
\usepackage{multirow}
\usepackage{ragged2e}
\usepackage{booktabs}
\usepackage{hhline}
\usepackage{floatrow}
\newfloatcommand{capbtabbox}{table}[][\FBwidth]
\usepackage{booktabs}
\usepackage{caption} 
\usepackage{pifont}

\usepackage[accsupp]{axessibility}  

\usepackage[pagebackref,breaklinks,colorlinks,citecolor=eccvblue]{hyperref}

\usepackage{orcidlink}

\begin{document}

\title{FSMDet: Vision-guided feature diffusion for fully sparse 3D detector} 

\titlerunning{Abbreviated paper title}

\author{Tianran Liu\inst{1}\orcidlink{0000-0002-8238-3767} \and
Morteza Mousa Pasandi\inst{1}\orcidlink{0009-0005-6801-4446} \and
Robert Laganiere\inst{1}\orcidlink{0000-0001-9475-8151}}

\authorrunning{Tianran et al.}

\institute{VIVA Lab, Department of EECS, University of Ottawa, Ontario, Canada \\
\email{\{tliu157, mmous029, laganier\}@uottawa.ca}}

\maketitle

\begin{abstract}
Fully sparse 3D detection has attracted an increasing interest in the recent years. However, the sparsity of the features in these frameworks challenges the generation of proposals because of the limited diffusion process. In addition, the quest for efficiency has
led to only few work on vision-assisted fully sparse models. In this
paper, we propose FSMDet (Fully Sparse Multi-modal Detection), which use visual information to guide the LiDAR feature diffusion process while still maintaining the efficiency of the pipeline. Specifically, most of fully sparse works focus on complex customized center fusion diffusion/regression operators. However, we observed that if the adequate object completion is performed, even the simplest interpolation operator leads to satisfactory results. Inspired by this observation, we split the vision-guided diffusion process into two modules: a Shape Recover Layer (SRLayer) and a Self Diffusion Layer (SDLayer). The former uses RGB information to recover the shape of the visible part of an object, and the latter uses a visual prior to further spread the features to the center region. Experiments demonstrate that our approach successfully improves the performance of previous fully sparse models that use  LiDAR only and reaches SOTA performance in multimodal models. At the same time, thanks to the sparse architecture, our method can be up to  5 times more efficient than previous SOTA methods in the inference process.

  \keywords{RGB-LiDAR Fusion \and Fully sparse 3D detection \and Shape recovery}
\end{abstract}

\section{Introduction}
\label{sec:intro}

The 3D detection of objects using LiDAR has made great progress in the past few years.  However, the vast majority of dense detector, whether they be anchor-based \cite{shi2023pv,deng2021voxel,zhou2018voxelnet,shi2020pv,lang2019pointpillars,mao2021voxel,li2023pillarnextrethinkingnetworkdesigns,chen2022focal}  or anchor-free \cite{zhou2022centerformercenterbasedtransformer3d,yin2021center} works on dense Bird's Eye View (BEV) representations which are costly to process. For instance, for every doubling of the detection range, the computational cost will increase by 2-4 times\cite{fan2022fullysparse3dobject}. To overcome this issue, fully sparse detectors have witnessed an increased interest in recent years. Different from point-based methods\cite{shi2019pointrcnn3dobjectproposal,qi2017pointnetdeephierarchicalfeature,qi2019deep,shi2020pointgnngraphneuralnetwork}, which perform time-consuming point neighborhood aggregation, fully sparse models tend to introduce a voxel or a point-wise segmentation to significantly reduce the computation cost early in the network.

However, sparse models come also with their issues. Two main questions in fully sparse detectors are center feature missing and image signal fusion. Specifically, for the detectors that adopt a dense detection head on the BEV map, the normal pipeline first feeds the voxelized lidar signal to backbone, which always consists of a stack of submanifold sparse convolution layers\cite{3DSemanticSegmentationWithSubmanifoldSparseConvNet}. However, this step won't change the occupancy of voxels. After the backbone, the z-axis(height) will be flattened to obtain the BEV map, which will still be sparse. Normally, the 2D CNN will process the BEV maps to diffuse the features to the empty voxels and then the detection head can generate the proposals for every non-empty voxel. The feature located in the voxel at the object's center will be the main candidate to regress the predicted box of the object. However, this method won't solve the issues created by missing voxel center feature in a fully sparse pipeline. Many fully sparse pipelines propose modules that vote the center of objects directly\cite{fan2022fullysparse3dobject} or use the voted center as virtual voxels\cite{fan2023fsd} to further aggregate features. However, this is a complex strategy that requires several hand-craft parameter settings which makes the model far from satisfactory in terms of deployment and speed of inference. It is worth noting that the recently proposed SAFDNet\cite{zhang2024safdnet} aims at achieving to achieve this goal through a foreground-only feature diffusion approach, but this parameter-free operation makes it difficult to ensure not only the accuracy of the diffusion but also the quality of the diffused features.

\begin{figure}[t]
\begin{center}
   \includegraphics[width=0.9\linewidth]{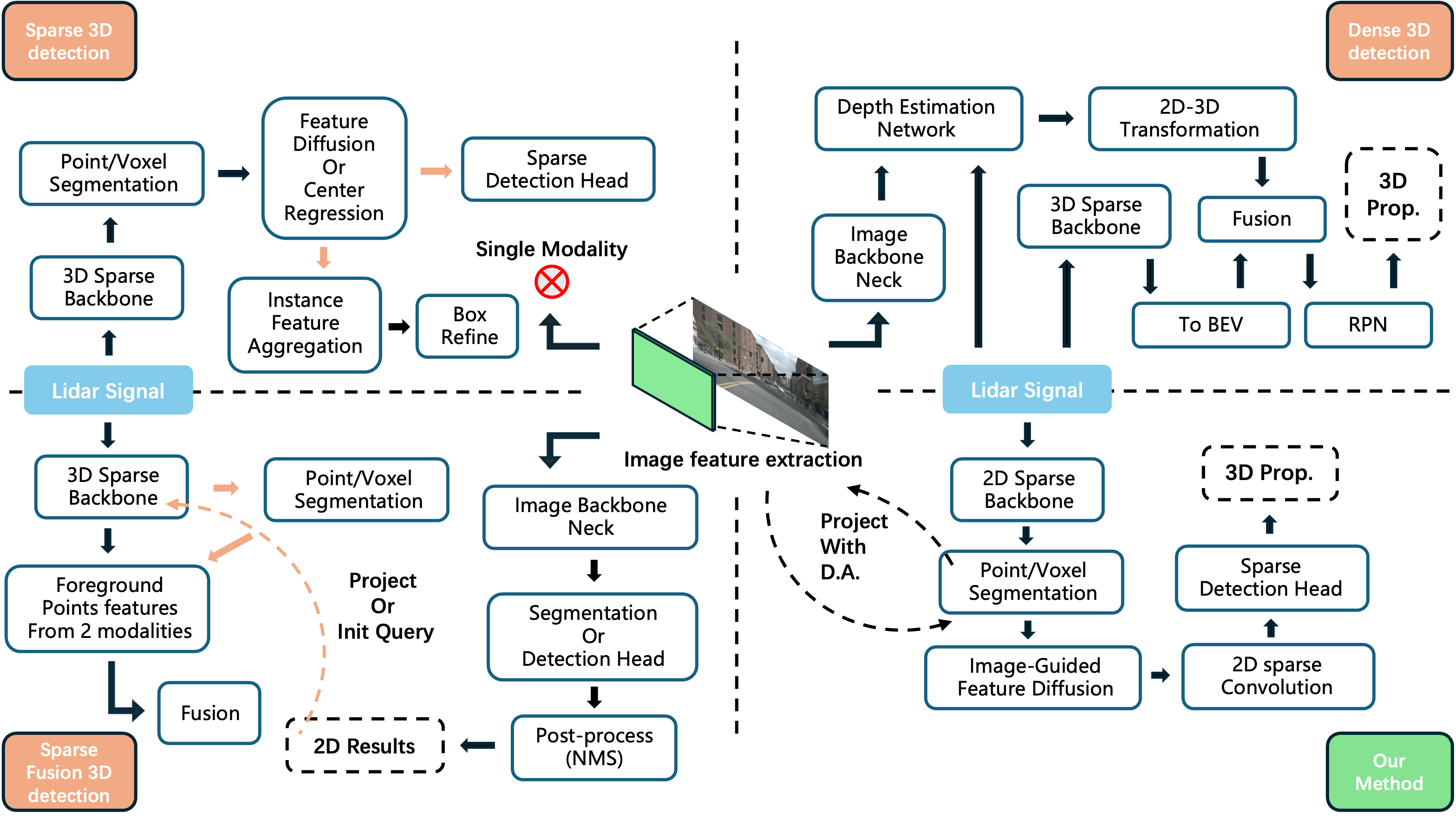}
\end{center}
\caption{A Vision-Centric view for LiDAR-RGB fusion in a fully sparse framework. Our proposed FSMDet integrates image features without any parameters for 2D tasks. The feature diffusion is guided by RGB features. Prop. and D.A. stand for proposals and deformable attention. The orange arrows represent the optional steps for models in the same categories.
\label{fig:pipeline}}
\end{figure}

In this work, we demonstrate that vision information can be used to guide the feature diffusion in a 3D detection framework. Only a few works\cite{ERABATI2024127814,xie2023sparsefusion,li2024fully} tried to introduce the RGB features into a fully sparse pipeline. However, most of them need a 2D segmentation or detection head to localize the objects from the image signal. These modules which include RPN and NMS operations limit the overall inference speed of the 3D detection framework. In other words, 2D detection is only intended to be used as an assist to the subsequent 3D detection results. Such a simple merged network tends to introduce a large amount of parameter redundancy. This inefficiency in RGB fusion approaches in fully sparse networks led us to revisit how to introduce RGB information while maintaining architectural sparsity. This paper introduces the FSMDet network (Fully Sparse Multi-modal Detection), a straightforward and efficient multi-modal 3D detector, that only keeps the necessary parameters for the image-branch and retains the sparsity of the structure.

As shown in Fig \ref{fig:pipeline}, our proposed method is as follows: the voxelized LiDAR features  interact with RGB features through deformable attention operations to select the foreground RGB features. The foreground voxel features are diffused in 2 ways: using the shape-recovery layer (SRLayer) and the self-diffusion layer (SFLayer). Specifically, considering that the diffusion process changes the occupancy, we believe that the process of an ideal feature diffusion should essentially be like shape completion. From an idealized test, we experimentally proved that for a fully sparse model, even if there is no specific design for the center feature diffusion, we can still get nearly 100\% accuracy with the completed objects. This conclusion can also be extended to most point-based methods. However, since the full shape of objects is not always available in the image, it is not a trivial task to make a network predicting the full shape of objects. Therefore, we further tested the visible part completion with color information. The result shows that this is even better for small objects with almost the same performance in other categories compared with the full-shape competition.

The experiment shows that a shape recovery operation at the early stage of the network is essential for accuracy. Inspired by this observation, the SRLayer is used at an early stage of the network to recover the shape of the visible part with features from the camera and the LiDAR modalities. The self-diffusion layer (SDLayer) is performed before the sparse detection head, which expands the feature at the boundary of objects to the center, thus solving the center feature estimation problem mentioned before. 

Our proposed FSMDet network efficiently and effectively introduces visual information into a fully sparse detector: on a desktop GPU, without any CUDA kernel or deployment optimization, our method achieve a competitive accuracy comparing with SOTA fusion-based solutions with up to 5.3x speedup.

\section{Related Work}

\textbf{Fully sparse 3D detection.} A premise that has sometimes been overlooked in past 3D detection methods\cite{deng2021voxel,shi2020pv,liang2022bevfusion} is the different importance of differentiating foreground objects from the background which significant impact computing resources. FSD \cite{fan2022fully} first proposed a solution with a segmentation head to distinguish objects from background signals and a voting network to regress the center of objects. Subsequently FSDv2\cite{fan2023fsd} improves the performance by introducing virtual voxels and a multi-stage pipeline. Later, the FSD++\cite{fan2023super} utilized multi-frame information to improve the segmentation process. Considering the complexity of the pipeline, VoxelNeXt \cite{chen2023voxelnext} adopt a relatively simple structure with extra down-sampling steps, which get better results with lower latency. The SAFDNet\cite{zhang2024safdnet} proposed another strategy that used a 2D sparse convolution layer named AFD to diffuse the foreground lidar signal. Although most of the works achieved the SOTA performance, none of the mentioned works introduced multi-modal signal analysis into the fully sparse 3D detection pipeline.
\\
\\
\textbf{Object Completion in 3D detection.} 
The concept of object completion starts from lidar-based 3D detection: With a sub-network recovering the missing signal of distant or occluded objects, the performance of 3D detection can be boosted. Specifically, BtcDet\cite{xu2021behind} used cylinder coordination to voxelise the space and then try to recover the missing signal by predicting the occupancy of voxels. SPG\cite{xu2021spg} tries to use an unsupervised method to expand the foreground voxels. Both PG-RCNN\cite{koo2023pg} and PC-RGNN\cite{zhang2021pc} tried to densify the foreground region after proposing the region in the first stage. Sparse2Dense \cite{wang2022sparse2dense} proposed a new way to complete the object in the hidden space by an explicit optimization. This same idea is behind many lidar-RGB fusion solutions. SFDNet\cite{wu2022sparse} and VirConv\cite{VirConv} introduce dense pseudo points to complete the objects and the latter further proposed different modules to remove possible noise. Distinct from these 2 works, WYSIWYD\cite{liu2023detectbetterobjectdensification} introduces a mesh deformation method that only completes the depth of foreground parts visible in the RGB image. 

\section{FSMDet}
\label{sec:fsmdet}

\subsection{Preliminary}
\label{sec:pre}
Previous works in fully sparse detection \cite{fan2023fsd,zhang2024safdnet,fan2022fullysparse3dobject} attribute its performance flaw to center feature aggregation, and then proposed different methods to improve the performance of this module. In this section, we introduce a series of ideal experiments to show that another important factor that affects the performance of a sparse 3D detector is the density of the foreground objects point cloud representation. In this first experiment, our objective is to test the performance of sparse models with plain center feature aggregation modules under different lidar inputs. Note that when choosing the models to be tested, it is important to select ones that do not change the center voxels' occupancy. This way, the change in performance can be attributed to the different lidar inputs.

\begin{figure}[t]
\begin{center}
   \includegraphics[width=\linewidth]{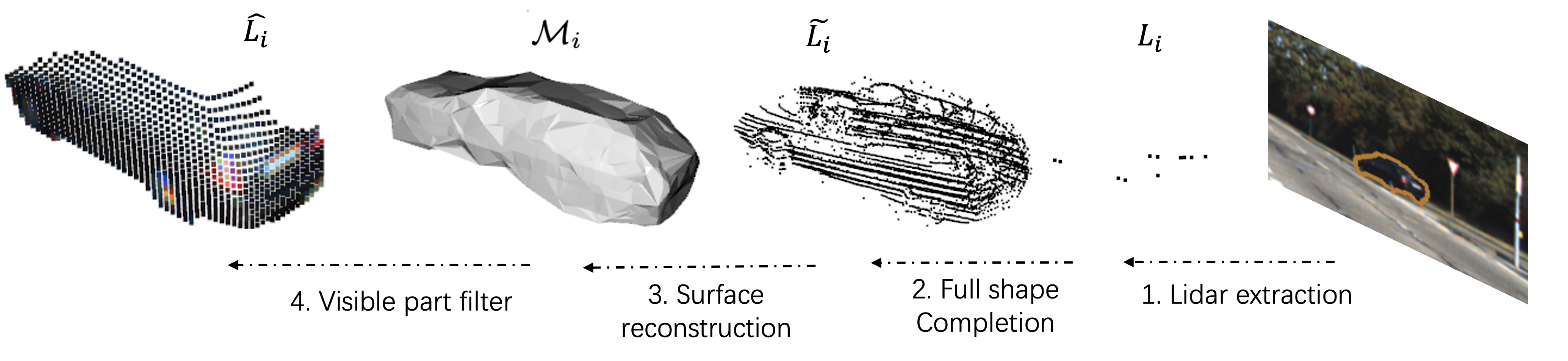}
\end{center}
\caption{From sparse lidar signal to visible part ground truth: we first obtain the full shape objects and then reconstruct the surface. With a ray-cast model, we can paint the color information onto the object's surface and filter out the non-visible part. 
\label{fig:fsvp}}
\end{figure}

With a 3D ground truth box $B_i$ of object i, we can obtain the lidar point set in this box, denoted by $L_i$. Using the full shape(FS) object point completion methods described in \cite{xu2021behind,liu2023detectbetterobjectdensification}, we can obtain a densified object, denoted by $\widetilde{L_{i}}$, as shown in Fig \ref{fig:fsvp}. Our experiment, under ideal condition, consists then in completing the full shape of all ground-truth foreground objects in a scene. For this experiments, we use 3 classic models: IA-SSD\cite{zhang2022pointsequallearninghighly}, Point-RCNN\cite{shi2019pointrcnn3dobjectproposal}, and FSD\cite{fan2022fullysparse3dobject}. As shown in the FS line of Table \ref{tab: upper}, even for these models without a modern center feature aggregation module, good performance can be obtained if the objects are completed. This suggests that even the most basic center regression method is sufficient for well-completed objects to give us good results in 3D detection.

However, although full-shape completion can lead to a good results, it is still a non-trivial task to recover the full shape either from the image or the lidar signal. Therefore, we further use the obtained $\widetilde{L_{i}}$ to generate the visible part of objects, denoted by $\hat{L_{i}}$. Specifically, for every single $L_i$ in the dataset, a unique 2D instance $I_i$ on the image can be identified. First, we adopted a surface reconstruction from the Possion\cite{kazhdan2006poisson} method, and then placed the obtained hull(reconstructed surface), denoted by $\mathcal{M}_i$ in 3D space. Based on the known camera intrinsic matrix, extrinsic matrix, and the pixel 2D coordinates of pixels in the region $I_i$, a ray casting model can be established. Note that since the rays from the pixel on the boundary sometimes miss the object, a volume expansion coefficient $\delta$ is used here to make sure there is always a bijection from the pixel set to the depth set. This process is illustrated in Figure \ref{fig:proj}.

\begin{table}[t]
    \centering
    \setlength{\tabcolsep}{0.3mm}{
    \begin{tabular}{cccccccc|cccccc}
    \toprule[0.28mm]
    \multirow{2}{*}{\centering Models} & \multirow{2}{*}{\makecell{Comp}} & \multicolumn{3}{c}{Car 3D $AP_{R40}$} & \multicolumn{3}{c}{Car BEV $AP_{R40}$} & \multicolumn{3}{c}{Ped 3D $AP_{R40}$} & \multicolumn{3}{c}{Ped BEV $AP_{R40}$}\cr 
			\cmidrule[0.25mm](lr){3-5}\cmidrule[0.25mm](lr){6-8}\cmidrule[0.25mm](lr){9-11}\cmidrule[0.25mm](lr){12-14}
			 & &Easy&Mod.&Hard&Easy&Mod.&Hard&Easy&Mod.&Hard&Easy&Mod.&Hard\cr
			\cmidrule[0.25mm](lr){1-14}
    	\cmidrule[0.25mm](lr){1-14}
     \multirow{2}{*}{\makecell{PointRCNN}} & VP & 99.9 &99.6&97.2&99.9 &99.6&97.2&85.3&77.9&86.2&68.1&80.9&71.2\cr 
     & FS& 97.2 &97.2&97.1&97.2 &97.2&97.1&77.8&72.8&70.3&80.3&74.8&71.1 \\
    \cmidrule[0.25mm](lr){1-14}
    \multirow{2}{*}{\makecell{IA-SSD}} & VP & 99.9 &99.7&99.4&99.9&99.3&99.3&75.5&71.3&66.4&77.5&75.1&70.3\cr
    & FS & 99.7&99.4&99.4&99.6&99.2&99.4&71.9&69.7&86.2&74.8&74.3&72.8\\
    \cmidrule[0.25mm](lr){1-14}
    \multirow{2}{*}{\makecell{FSD-V1}} & VP & 99.9&99.7&99.7& 99.9&99.7&99.7&87.2 &80.1 &72.1&89.9&82.7&73.6\cr & FS &99.9&99.9&99.9&99.9&99.9&99.9&81.7&77.9&71.1&83.3&77.1&72.9 \\
    \cmidrule[0.25mm](lr){1-14}
    \end{tabular}
    
    }
    \caption{The performance of different sparse models with completed objects on the KITTI validation set. FS and VP stand for full shape and Visible Part completion respectively. 
     }
    \label{tab: upper}
\end{table}

With the equation of line $T_c$ and $\mathcal{M}_i$, the distance traveled by a ray of light from point $c$ when it arrives on the mesh can be expressed as $\left| \delta  \mathcal{M}_i \cap T_c -c_i \right| $. Considering the slope of $T_C$, the depth of the point on the mesh hit by the light to the camera plane is $d_i$. Here $\cdot$ refers to scalar multiplication. 

\begin{equation}\label{dc}
	d_c = \left| \delta  \mathcal{M}_i \cap T_c -c_i \right| \cdot \left|\vec{x_c}\right|
\end{equation}

\begin{figure}[h]
\begin{center}
   \includegraphics[width=0.6\linewidth]{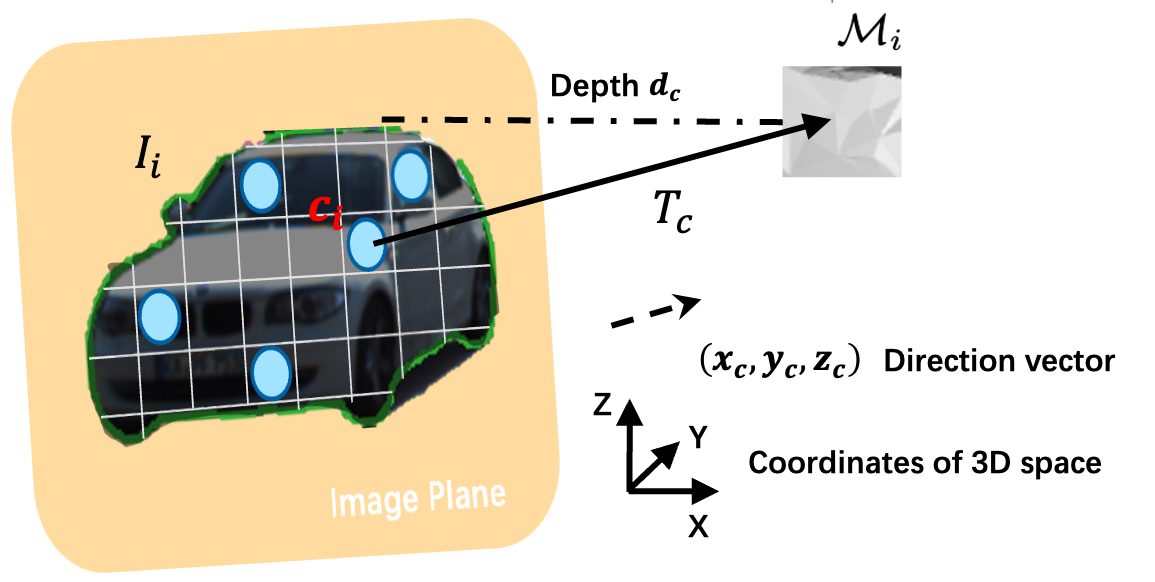}
   \label{fig:proj}
\end{center}
\caption{ The ray casting model for visible part ground truth generation. All calculations are performed in LiDAR(3D space) coordination. Here $c_i$ stand for the location of pixel c in 3D space.
\label{fig:proj}}
\end{figure}

After estimating the depth of every pixel in the region $I_i$, we get a dense visible surface of the object, denoted by $\hat{L_i}$, as shown in the left most sub-figure of Fig \ref{fig:fsvp}. Replacing the original $\widetilde{L_{i}}$ with $\hat{L_i}$, we repeated the same experiment on the selected models in VP lines of Table \ref{tab: upper}, show that even with only visible part completion, the upper boundary of these models is still high enough. A counter-intuitive fact that also needs to be mentioned is that VP is even more suitable for small objects such as pedestrians. 

During the training, the completion of objects causes a change of occupancy, or in other words, it diffuses the point features. This experiment shows that the detection results are satisfactory when the object shapes are adequately completed, even with a vanilla design for the center regression or center feature aggregation modules. This inspired us to split the feature diffusion into two steps: the shape recovery diffusion and the diffusion of features to the object center. Since only the visible part of objects is available on the image, here we use VP to supervise the first-step diffusion. 

\subsection{Overall structure}

\label{sec:overall}

In Section \ref{sec:pre}, we experimentally demonstrated that the visible part completion of the foreground object in the LiDAR signal improve the detection results despite the diffculty of center prediction. Based on this result, we introduce the shape recovery module, which diffuses the features to occupy near voxels, at an early layer of the network. Our experiments will also demonstrate that it's beneficial to do this step at an early stage.

\begin{figure}[t]
\begin{center}
   \includegraphics[width=\linewidth]{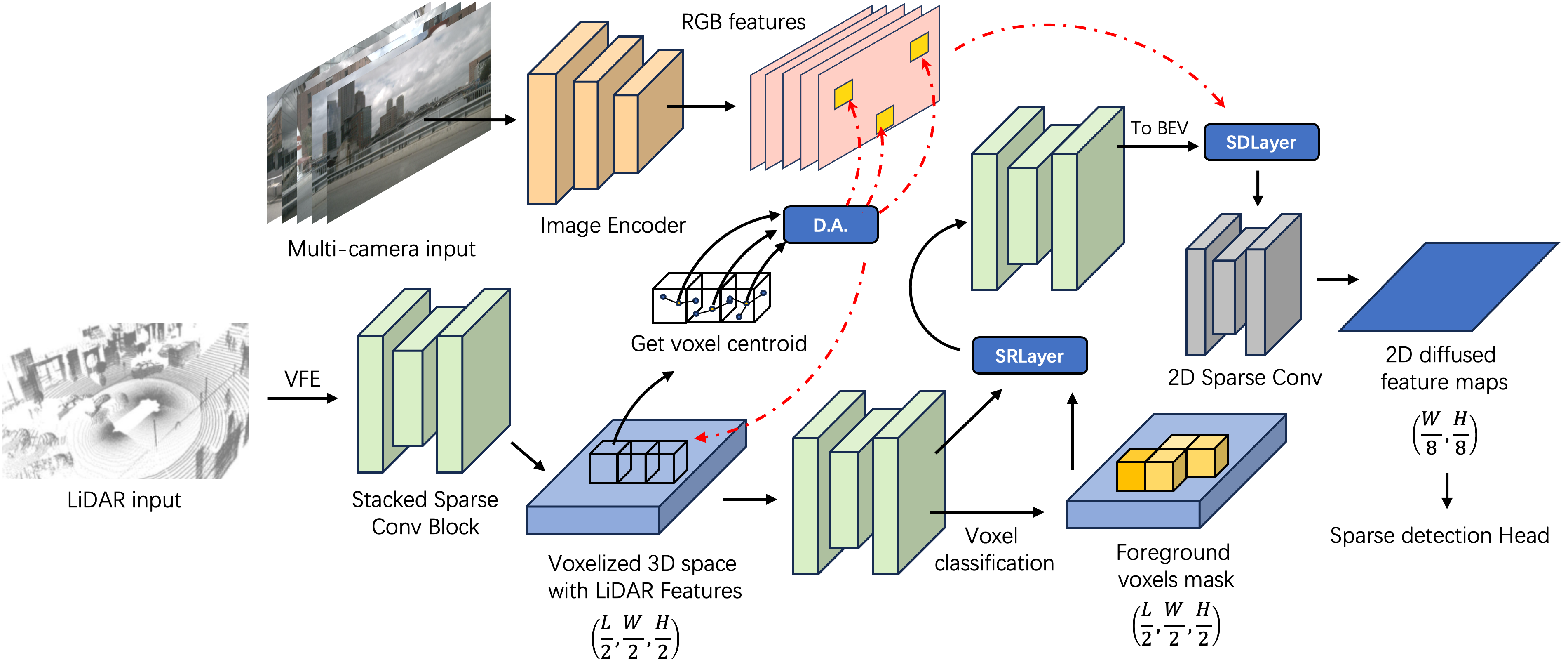}
\end{center}
\caption{Proposed pipeline: We complete the diffusion of features in two steps, the SRLayer uses RGB features to recover the visible part of the contours of objects while the SDLayer diffuses the features to the center of objects. D.A. stand for the deformable attention. The operation related to color features is marked in red.
\label{fig:main}}
\end{figure}

Specifically, as shown in Fig \ref{fig:main}, we use SRLayer (Shape Recover Layer) and SDLayer (Self Diffusion Layer) to do the two steps mentioned respectively. After the voxelization of the space, Submanifold sparse convolution\cite{graham2017submanifoldsparseconvolutionalnetworks} with residual design was used as the basic block to process the LiDAR data. As in work \cite{zhang2024safdnet}, we call these stacked sparse convolution block \textbf{SRB}(Sparse Residual Block). We choose $\frac{1}{4}$ of the original size to perform shape recovery and color features aggregation as a good balance between efficiency and performance. With the coordinates of points centroid in the different voxels and the calibration information, we fuse the voxel and image features with a deformable attention operator and replace the original feature. These augmented features will be further process with SRBs, and then a voxel classification head will be applied to get the foreground parts of the scene. The fused features and the foreground masks are used as the input of our SRLayer. SRLayers use 2D VP object completion as the supervision signal.

The output of SRLayers will go through the SRBs for further downsampling. Finally, the BEV transformation squeezes the feature map to get a 2D representation of the space, the SDLayer will be used to deliver the feature to the center. The  output size of the proposed cross-modality backbone will be $\frac{1}{8}$ of the width and length of the original input.

\subsection{Proposed Method}
\label{sec:prop}

\textbf{Deformable attention for feature fusion.} Let $V=\left\{ V_i, V_{f_{i}}, c_i \right\}_{i=1}^{|V|}$ represent all of the non-empty voxels in a sparse feature map, $V_i$ is the indices of the voxel i, $V_{f_{j}}$ stand for the features in the voxel and $c_i$ is the centroid of the LiDAR points in this voxel. With the camera intrinsic and extrinsic matrix, the 3D-2D transformation can be denoted as $\mathcal{P}$, to obtain the 2D-pixel coordinates $\textbf{p}_i$, as shown in equation \ref{eq:proj}.

\begin{equation}
\label{eq:proj}
\textbf{p}_i = \mathcal{P}(c_i)
\end{equation}

Normally, we will interpolate the image features to the original size and acquire the image feature for points in 2D coordinates. However, as pointed out in previous lidar segmentation work\cite{zhang2023lidarcamerapanopticsegmentationgeometryconsistent}, the calibration of different modalities is not always reliable. Therefore the correspondency between semantic information in down-sampled image feature maps and voxel centroid point is not guaranteed. For this reason, we introduce deformable attention in this step to allow the network to freely select features near $\textbf{p}_i$, following the design of LoGoNet\cite{li2023logonetaccurate3dobject} with a minor modification. 

Specifically, $V_{f_{i}}$ will be used as query of the attention calculation. The image features sampled from the pixels around $\textbf{p}_i$, denoted by $\hat{R}_i$, will be used as Key and Value. The image features located at $\textbf{p}_i$ and all of its 3-order neighbour will be noted as $R_{\textbf{p}_{i}}$ and $R_{\textbf{p}_{i}}^{3}$, with the image map itself denoted by $R$. With M attention head and K sample points, we can obtain $\hat{R}_i$ from Equation \ref{eq:da}.

\begin{equation}
    \label{eq:da}
    \begin{gathered}
    \Delta\textbf{p}_{mki} = MLP(MLP(R_{\textbf{p}_{i}}^{3}) \cdot V_{f_{i}}) \\
    R_{mi} = \mathcal{G}(R, \textbf{p}_i + \Delta\textbf{p}_{mki}) \\
    \hat{R}_i = \sum_{m=1}^M W_m\left[\sum_{k=1}^K A_{m i k} \cdot\left(W_m^{\prime} R_{mi}\right)\right]
    \end{gathered}
\end{equation}

The $\mathcal{G}(a,b)$ represents the operation of obtaining the features of a at location b. The $W_m^{\prime}$, $W_m$ stands for the trainable weight, and $A_{m i k}$ stands for attention weight. In the original design, the $\Delta\mathbf{p}_{mki}$ was only decided by the LiDAR feature $V_{f_{i}}$, however, here we add an interaction between $V_{f_{i}}$ and all features around $R_{\textbf{p}_{i}}$, which will help to identify the proper features to be aggregated. 
\\
\\
\textbf{Shape Recover Layer.} The augmented features will replace the original one and is sent to the next SRBs. Note that at this stage, we keep the original feature map size. The obtained map is then sent to a classification head to split the foreground voxels from the background information. Considering the large space difference between the various classes of objects, we set a category number $\mathcal{S}$ for the classification. Using $\mathcal{M}$ to represent the index of foreground voxels, the $\mathcal{M}$ and processed voxels $V^\prime = \left\{ V_i, V_{f_{i}}^\prime, c_i \right\}_{i=1}^{|V|}$will be sent to the SRLayer.

\begin{figure}[h]
\begin{center}
   \includegraphics[width=0.9\linewidth]{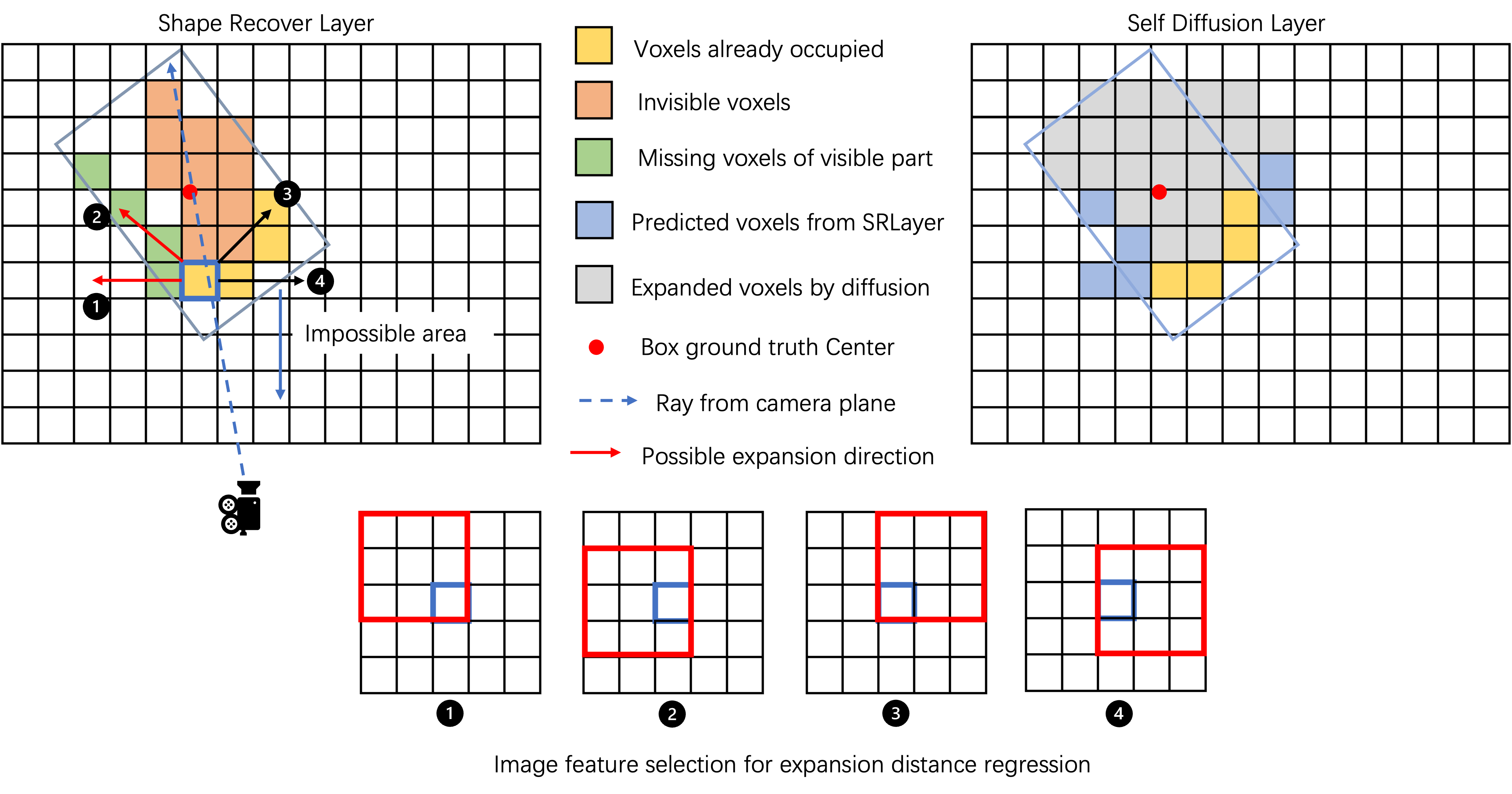}
\end{center}
\caption{Illustration of Shape Recover Layer and Self Diffusion Layer. We use the sparse signal to recover the shape of objects. After Shape recovery, we expand the boundary voxels along the ray direction, the center point will be naturally located in this area.
\label{fig:SR&SD}}
\end{figure}

With the foreground voxels that are denoted by $V^\prime_{o}=\left\{V_i, V_{f_i}^\prime,c_i | V_i \in \mathcal{M}\right\}_{i=1}^{|\mathbf{M}|}$, we first need to calculate the possible direction of diffusion. A possible direction should ensure that the voxel in that direction is visible on the image and secondly that the next two voxels in that direction are not occupied. As shown in the left part of Fig \ref{fig:SR&SD}, all voxels in $V_o^{\prime}$ will have 4 candidate directions to diffuse, but here only 2 of them are possible for diffusion (marked with red). Specifically, for a specific voxel in $V_o^\prime$, denoted by $\left\{V_t, V_{f_t}^\prime,c_t\right\}$, which can be projected to a pixel located at $\textbf{p}_t$, we use $V_{\varnothing}$ to denote all invisible voxels. The possible diffusion direction $J_t$ can be described in Equation \ref{eq:direct}. Here $V_j$ is the nearest non-empty voxels on direction $J_t$.
\begin{equation}
    \label{eq:direct}
    J_t = \left\{J_{ti}|J_{ti} \cap V_{\varnothing} , \left\|V_t-V_j\right\|<2\right\}_{i=1}^4
\end{equation}

For every $J_{ti}$, we use $V_{f_t}^\prime$ and the direction-specific image features (as shown in the bottom of Figure \ref{fig:SR&SD}), denoted by $R_{\textbf{p}_t}^i$, to regress the expand distance $d_t$. For the empty voxel in range $d_t$ alone the direction $J_t$, denoted by $\left\{V_{t+d}, V_{f_{t+d}}^\prime, c_{t+d}\right\}$, the $V_{f_{t+d}}^\prime$ will be obtain by equation \ref{eq:feature}.
\begin{equation}
    \label{eq:feature}
    V_{f_{t+d}}^\prime = MLP(\bigcup_{\mathbf{N}(p_{t+d}, d)}R_{\textbf{p}_{t+d}}^i)
\end{equation}
Here $\mathbf{N}(x, l)$ and $\bigcup$ stand for the $l$-order neighborhood of pixel $x$ and the concatenation operation of the neighboring features. With the new occupied voxels, the $V^\prime$ will be sent to the next SRB to process the features further.

The Shape Recover Layer needs supervision from the generated VP ground truth. Considering the computational complexity, we only apply the supervision on the BEV plane. Specifically, after voxelizing the VP ground truth, we can get the ground truth for the shape recovery, denoted by $V_{gt}$. Using $\hat{V}$ to represent the score of predicted recovery voxels, the loss can be written as Equation \ref{eq:loss}.
\begin{equation}
    \label{eq:loss}
    \mathcal{L} = \sum_{i=1}^\mathbf{S} \alpha_j \sum_{j=1}^3 focal\_loss(\hat{V_{ij}}, V_{gt})
\end{equation}
Note that the boundary voxels only make up a small portion. Here we use a multi-class focal loss to calculate if the predicted occupied voxel cannot cover the boundary ground truth. If it is located in the 2D box, it won't be treated as negative samples. Here we further split the voxel predicted for different size objects into 3 categories: the predicted voxels located at the boundary correctly, the predicted voxels not situated at the boundary but in the box, and the others. Specific $\alpha_j$ will be used to scale these different cases here. 
\\
\\
\textbf{Self Diffusion Layer.} Before being sent to the detection head, the sparse signal needs to be further diffused to the center of objects. Note that since we have identified the invisible voxels $V_\varnothing$, we can directly diffuse the features along the ray across the occupied voxels. As we have shown in the right part of Figure \ref{fig:SR&SD}, in most cases, the center can be naturally covered. Considering the size of different objects, the scope parameter $\sigma_0$ of this diffusion depends on the result of the classification head. Following SAFDNet\cite{zhang2024safdnet}, we also adopt an adaptive method in which for larger objects, the diffusion distance alone ray casting will be larger than that of small objects. However, when the objects are occluded, the diffusion along the ray-casting direction will miss the center. Therefore, considering the size of voxels here, we also set a parameter $\sigma_1$ to control the diffusion of features perpendicular to the direction of the ray casting.

\section{Experiments}
\subsection{Dataset and Metrics}
We conducted experiments on the nuScenes \cite{caesar2020nuscenesmultimodaldatasetautonomous} dataset for a comparison. The nuScenes dataset was collected with 6 cameras, a radar sensor, and a 32-beam LiDAR sensor under different weather and illumination conditions. The model proposed was evaluated by the official evaluation script with NDS and MAP as index.

\subsection{Implementation Details}

We used the OpenPCDet\cite{openpcdet2020} as the framework to implement the mentioned design. All experiments were performed on RTX3090, we used 8 as batch size on 4 GPUs for training and 1 batch size for inference speed test. Training in all experiments lasted for 20 epochs. During training, in addition to random rotation and translation of the point cloud signals and random cropping and normalization of the image signals, we change the GT-sampling commonly used in the lidar-based model to the multimodal GT-sampling proposed in PointAugmenting\cite{Wang_2021_CVPR}. We use ResNet\cite{he2016deep} pre-trained on NuImage as the image backbone. The volume expansion parameter
$\delta$ was set to 1.15 to ensure all rays from every pixel wouldn't miss the object. $d$ in Equation \ref{eq:feature} was set to 2. Considering the size of different objects occupied, we set $\sigma_0=6$ and $\sigma_1=4$. We set the $\alpha_0=\alpha_1=0.5$, $\alpha_2=1$ to scale the loss mentioned in Equation \ref{eq:loss}.

\subsection{Main results}

\begin{table}[h]
	\centering
	\footnotesize
	\setlength{\tabcolsep}{1.5mm}{
		\begin{tabular}{cccccccc}
			\toprule[0.25mm]
			\toprule[0.25mm]
			Methods& mAP$\uparrow$ & NDS$\uparrow$ & mATE$\downarrow$ & mASE$\downarrow$ & mAOE$\downarrow$ & mAVE$\downarrow$ & mAAE$\downarrow$ \cr
			
			\cmidrule[0.25mm](lr){1-8}

            \cmidrule[0.25mm](lr){1-8}
            FSDv2 & 65.4 & 70.8 & 0.270 & 0.248 & 0.272 & 0.207 & 0.187 \\
            
            SAFDNet & 68.3 & 72.3 & 0.251 & 0.242 & 0.311 & 0.258 & 0.127 \\

            SparseFusion & 71.0 & 73.1 & 0.277 & 0.247 & 0.270 & 0.253 & 0.188 \\
            MSMDFusion & 69.2 & 72.0 & 0.283 & 0.254 & 0.282 & 0.252 & 0.185 \\
            BEVFusion & 67.3 & 70.7 & 0.286 & 0.255 & 0.313 & 0.251 & 0.186 \\
            FSMDet & 70.0 & 72.2 & 0.285 & 0.251 & 0.296& 0.254 & 0.179 \\
			\bottomrule[0.25mm]
			\bottomrule[0.25mm]
			
	\end{tabular}}
	\vspace{2mm}
	\caption{Performance Comparsion with SOTA models on \textbf{nuScences validation set}. The best results are shown in bold. $\uparrow$ higher is better, $\downarrow$ lower is better.}
	\label{ResultNu}
\end{table}

We compared the inference speed with recent SOTA models. Since we did not need any parameters for tasks on the image plane, our proposed method takes only $25\%$(as a minimum) of the time the SOTA solutions use to take while maintaining a decent detection performance. As shown in Table \ref{table:infer}, our proposed method obtained 70.1 mAP and 72.1 NDS in nuScenes Dataset. This performance is only $2.4\%$ to $3.8\%$ less than the SOTA but we only used $18\%$ to $47\%$ of the inference time that is needed for SOTA models. 

\begin{table}[h]
\setlength{\tabcolsep}{1.5mm}{
    \begin{tabular}{ccccccc}
        \toprule
        Model     & Modality   & Sparsity & Image Result & FPS & mAP & NDS \\
        \midrule
        SAFDNet & LiDAR & \checkmark & N/A & 15.7 & 68.3 & 72.3   \\
        
        SparseFusion & LiDAR+RGB & \checkmark & \checkmark & 5.3 & 71.0 & 73.1 \\

        MSMDFusion & LiDAR+RGB & \ding{53} & \ding{53} &2.1& 71.4&73.9       \\
        BEVFusion & LiDAR+RGB & \ding{53} & \ding{53}& 8.1 & 67.3 & 70.7  \\
        \midrule
        FSMDet & LiDAR+RGB & \checkmark & \ding{53} & 11.2 & 70.0&72.2 \\
        \bottomrule
    \end{tabular}}
    \caption{Efficiency comparison, our proposed model achieves a balance point between inference speed and performance.}
    \label{table:infer}
\end{table}

\subsection{Ablation Study}
\label{sec:abla}

In Table \ref{table:comp}, we break down the performance boost in the proposed model. Here the D.A Color and Proj Color stand for the two different methods to introduce RGB color features: the deformable attention or direct projection. As shown in the table, the parameter-based solution benefits from the contribution of these models. In experiments on Table \ref{table:comp}, if we do not apply the SDLayer, the ADF in SAFDNet\cite{zhang2024safdnet} is used.

\begin{table}[h]
    \centering
    \setlength{\tabcolsep}{1.5mm}
    {
    \begin{tabular}{ccccccc}
        \toprule
        Models & D.A Color & Proj Color & SRLayer & SDLayer & mAP & NDS \\
        \midrule
        FSMDet & \checkmark & - & - & - & 67.9 & 70.4 \\
        FSMDet & - & \checkmark & - & - & 67.6 & 70.8 \\
        FSMDet & \checkmark & - & \checkmark & - & 69.8 & 71.5 \\
        FSMDet & \checkmark & - & \checkmark & \checkmark & 70.0 & 72.2 \\
        \bottomrule
    \end{tabular}
    }
    \caption{The verification of the effect of each module on the model on the final accuracy}
    \label{table:comp}
\end{table}

Here in Table \ref{table:sr-layer}, we compare the effect of different positions of the SRLayer on accuracy. If we insert the shape recovery operation too early, the shallow representation of LiDAR features can barely help the expansion regress. On the other hand, if the SRLayer is too close to the SDLayer, the feature of the newly occupied layer also cannot aggregate the nearby features very well.

\begin{table}[h]
    \centering
    \setlength{\tabcolsep}{1.5mm}
    {
    \begin{tabular}{ccccccc}
        \toprule
        Modules & Block-1 & Block-2& Block-3 & Block-4 & mAP & NDS \\
        \midrule
        SRLayer & \checkmark & - & - & - & 61.2 & 63.6 \\
        SRLayer & - & \checkmark & - & - & 70.0 & 72.2 \\
        SRLayer & - & - & \checkmark & - & 68.7 & 70.9 \\
        SRLayer & - & - & - & \checkmark & 63.1 & 67.2 \\
        \bottomrule
    \end{tabular}
    }
    \caption{Different stages to insert the Shape Recover Layer}
    \label{table:sr-layer}
\end{table}

\label{sec:abla}
\section{Conclusion} 
In this work, we proposed a cross-modality method for the RGB-LiDAR fusion in a fully sparse framework. Starting from the idealized experiment which reveals that the core of center regression for detection models should be the visible part completion, we designed the SRLayer to achieve this step with the guidance from the image features. This way, the diffusion alone of the ray from the image plane can naturally cover the center of objects in most cases. As the experiments prove the potential of our method, we hope the FSMDet can promote the research in data fusion for fully sparse models.

\clearpage  

%
%
\bibliographystyle{splncs04}
\bibliography{main}

\begin{thebibliography}{10}
\providecommand{\url}[1]{\texttt{#1}}
\providecommand{\urlprefix}{URL }
\providecommand{\doi}[1]{https://doi.org/#1}

\bibitem{caesar2020nuscenesmultimodaldatasetautonomous}
Caesar, H., Bankiti, V., Lang, A.H., Vora, S., Liong, V.E., Xu, Q., Krishnan,
  A., Pan, Y., Baldan, G., Beijbom, O.: nuscenes: A multimodal dataset for
  autonomous driving (2020), \url{https://arxiv.org/abs/1903.11027}

\bibitem{chen2022focal}
Chen, Y., Li, Y., Zhang, X., Sun, J., Jia, J.: Focal sparse convolutional
  networks for 3d object detection. In: Proceedings of the IEEE/CVF Conference
  on Computer Vision and Pattern Recognition. pp. 5428--5437 (2022)

\bibitem{chen2023voxelnext}
Chen, Y., Liu, J., Zhang, X., Qi, X., Jia, J.: Voxelnext: Fully sparse voxelnet
  for 3d object detection and tracking. In: Proceedings of the IEEE/CVF
  Conference on Computer Vision and Pattern Recognition. pp. 21674--21683
  (2023)

\bibitem{deng2021voxel}
Deng, J., Shi, S., Li, P., Zhou, W., Zhang, Y., Li, H.: Voxel r-cnn: Towards
  high performance voxel-based 3d object detection. In: Proceedings of the AAAI
  conference on artificial intelligence. vol.~35, pp. 1201--1209 (2021)

\bibitem{ERABATI2024127814}
Erabati, G.K., Araujo, H.: Srfdet3d: Sparse region fusion based 3d object
  detection. Neurocomputing  \textbf{593},  127814 (2024).
  \doi{https://doi.org/10.1016/j.neucom.2024.127814},
  \url{https://www.sciencedirect.com/science/article/pii/S092523122400585X}

\bibitem{fan2022fully}
Fan, L., Wang, F., Wang, N., Zhang, Z.X.: Fully sparse 3d object detection.
  Advances in Neural Information Processing Systems  \textbf{35},  351--363
  (2022)

\bibitem{fan2022fullysparse3dobject}
Fan, L., Wang, F., Wang, N., Zhang, Z.: Fully sparse 3d object detection
  (2022), \url{https://arxiv.org/abs/2207.10035}

\bibitem{fan2023fsd}
Fan, L., Wang, F., Wang, N., Zhang, Z.: Fsd v2: Improving fully sparse 3d
  object detection with virtual voxels. arXiv preprint arXiv:2308.03755  (2023)

\bibitem{fan2023super}
Fan, L., Yang, Y., Wang, F., Wang, N., Zhang, Z.: Super sparse 3d object
  detection. IEEE transactions on pattern analysis and machine intelligence
  \textbf{45}(10),  12490--12505 (2023)

\bibitem{3DSemanticSegmentationWithSubmanifoldSparseConvNet}
Graham, B., Engelcke, M., van~der Maaten, L.: 3d semantic segmentation with
  submanifold sparse convolutional networks. CVPR  (2018)

\bibitem{graham2017submanifoldsparseconvolutionalnetworks}
Graham, B., van~der Maaten, L.: Submanifold sparse convolutional networks
  (2017), \url{https://arxiv.org/abs/1706.01307}

\bibitem{he2016deep}
He, K., Zhang, X., Ren, S., Sun, J.: Deep residual learning for image
  recognition. In: Proceedings of the IEEE conference on computer vision and
  pattern recognition. pp. 770--778 (2016)

\bibitem{kazhdan2006poisson}
Kazhdan, M., Bolitho, M., Hoppe, H.: Poisson surface reconstruction. In:
  Proceedings of the fourth Eurographics symposium on Geometry processing.
  vol.~7 (2006)

\bibitem{koo2023pg}
Koo, I., Lee, I., Kim, S.H., Kim, H.S., Jeon, W.J., Kim, C.: Pg-rcnn: Semantic
  surface point generation for 3d object detection. In: Proceedings of the
  IEEE/CVF International Conference on Computer Vision. pp. 18142--18151 (2023)

\bibitem{lang2019pointpillars}
Lang, A.H., Vora, S., Caesar, H., Zhou, L., Yang, J., Beijbom, O.:
  Pointpillars: Fast encoders for object detection from point clouds. In:
  Proceedings of the IEEE/CVF conference on computer vision and pattern
  recognition. pp. 12697--12705 (2019)

\bibitem{li2023pillarnextrethinkingnetworkdesigns}
Li, J., Luo, C., Yang, X.: Pillarnext: Rethinking network designs for 3d object
  detection in lidar point clouds (2023),
  \url{https://arxiv.org/abs/2305.04925}

\bibitem{li2023logonetaccurate3dobject}
Li, X., Ma, T., Hou, Y., Shi, B., Yang, Y., Liu, Y., Wu, X., Chen, Q., Li, Y.,
  Qiao, Y., He, L.: Logonet: Towards accurate 3d object detection with
  local-to-global cross-modal fusion (2023),
  \url{https://arxiv.org/abs/2303.03595}

\bibitem{li2024fully}
Li, Y., Fan, L., Liu, Y., Huang, Z., Chen, Y., Wang, N., Zhang, Z.: Fully
  sparse fusion for 3d object detection. IEEE Transactions on Pattern Analysis
  and Machine Intelligence  (2024)

\bibitem{liang2022bevfusion}
Liang, T., Xie, H., Yu, K., Xia, Z., Lin, Z., Wang, Y., Tang, T., Wang, B.,
  Tang, Z.: Bevfusion: A simple and robust lidar-camera fusion framework.
  Advances in Neural Information Processing Systems  \textbf{35},  10421--10434
  (2022)

\bibitem{liu2023detectbetterobjectdensification}
Liu, T., Zhang, Z., Pasandi, M.M., Laganiere, R.: What you see is what you
  detect: Towards better object densification in 3d detection (2023),
  \url{https://arxiv.org/abs/2310.17842}

\bibitem{mao2021voxel}
Mao, J., Xue, Y., Niu, M., Bai, H., Feng, J., Liang, X., Xu, H., Xu, C.: Voxel
  transformer for 3d object detection. In: Proceedings of the IEEE/CVF
  international conference on computer vision. pp. 3164--3173 (2021)

\bibitem{qi2019deep}
Qi, C.R., Litany, O., He, K., Guibas, L.J.: Deep hough voting for 3d object
  detection in point clouds. In: Proceedings of the IEEE International
  Conference on Computer Vision (2019)

\bibitem{qi2017pointnetdeephierarchicalfeature}
Qi, C.R., Yi, L., Su, H., Guibas, L.J.: Pointnet++: Deep hierarchical feature
  learning on point sets in a metric space (2017),
  \url{https://arxiv.org/abs/1706.02413}

\bibitem{shi2020pv}
Shi, S., Guo, C., Jiang, L., Wang, Z., Shi, J., Wang, X., Li, H.: Pv-rcnn:
  Point-voxel feature set abstraction for 3d object detection. In: Proceedings
  of the IEEE/CVF conference on computer vision and pattern recognition. pp.
  10529--10538 (2020)

\bibitem{shi2023pv}
Shi, S., Jiang, L., Deng, J., Wang, Z., Guo, C., Shi, J., Wang, X., Li, H.:
  Pv-rcnn++: Point-voxel feature set abstraction with local vector
  representation for 3d object detection. International Journal of Computer
  Vision  \textbf{131}(2),  531--551 (2023)

\bibitem{shi2019pointrcnn3dobjectproposal}
Shi, S., Wang, X., Li, H.: Pointrcnn: 3d object proposal generation and
  detection from point cloud (2019), \url{https://arxiv.org/abs/1812.04244}

\bibitem{shi2020pointgnngraphneuralnetwork}
Shi, W., Ragunathan, Rajkumar: Point-gnn: Graph neural network for 3d object
  detection in a point cloud (2020), \url{https://arxiv.org/abs/2003.01251}

\bibitem{openpcdet2020}
Team, O.D.: Openpcdet: An open-source toolbox for 3d object detection from
  point clouds. \url{https://github.com/open-mmlab/OpenPCDet} (2020)

\bibitem{Wang_2021_CVPR}
Wang, C., Ma, C., Zhu, M., Yang, X.: Pointaugmenting: Cross-modal augmentation
  for 3d object detection. In: Proceedings of the IEEE/CVF Conference on
  Computer Vision and Pattern Recognition (CVPR). pp. 11794--11803 (June 2021)

\bibitem{wang2022sparse2dense}
Wang, T., Hu, X., Liu, Z., Fu, C.W.: Sparse2dense: Learning to densify 3d
  features for 3d object detection. Advances in Neural Information Processing
  Systems  \textbf{35},  38533--38545 (2022)

\bibitem{VirConv}
Wu, H., Wen, C., Shi, S., Wang, C.: Virtual sparse convolution for multimodal
  3d object detection. In: CVPR (2023)

\bibitem{wu2022sparse}
Wu, X., Peng, L., Yang, H., Xie, L., Huang, C., Deng, C., Liu, H., Cai, D.:
  Sparse fuse dense: Towards high quality 3d detection with depth completion.
  In: Proceedings of the IEEE/CVF conference on computer vision and pattern
  recognition. pp. 5418--5427 (2022)

\bibitem{xie2023sparsefusion}
Xie, Y., Xu, C., Rakotosaona, M.J., Rim, P., Tombari, F., Keutzer, K.,
  Tomizuka, M., Zhan, W.: Sparsefusion: Fusing multi-modal sparse
  representations for multi-sensor 3d object detection. In: Proceedings of the
  IEEE/CVF International Conference on Computer Vision. pp. 17591--17602 (2023)

\bibitem{xu2021behind}
Xu, Q., Zhong, Y., Neumann, U.: Behind the curtain: Learning occluded shapes
  for 3d object detection. arXiv preprint arXiv:2112.02205  (2021)

\bibitem{xu2021spg}
Xu, Q., Zhou, Y., Wang, W., Qi, C.R., Anguelov, D.: Spg: Unsupervised domain
  adaptation for 3d object detection via semantic point generation. In:
  Proceedings of the IEEE/CVF International Conference on Computer Vision. pp.
  15446--15456 (2021)

\bibitem{yin2021center}
Yin, T., Zhou, X., Krahenbuhl, P.: Center-based 3d object detection and
  tracking. In: Proceedings of the IEEE/CVF conference on computer vision and
  pattern recognition. pp. 11784--11793 (2021)

\bibitem{zhang2024safdnet}
Zhang, G., Chen, J., Gao, G., Li, J., Liu, S., Hu, X.: {SAFDNet}: A simple and
  effective network for fully sparse 3d object detection (2024)

\bibitem{zhang2021pc}
Zhang, Y., Huang, D., Wang, Y.: Pc-rgnn: Point cloud completion and graph
  neural network for 3d object detection. In: Proceedings of the AAAI
  conference on artificial intelligence. vol.~35, pp. 3430--3437 (2021)

\bibitem{zhang2022pointsequallearninghighly}
Zhang, Y., Hu, Q., Xu, G., Ma, Y., Wan, J., Guo, Y.: Not all points are equal:
  Learning highly efficient point-based detectors for 3d lidar point clouds
  (2022), \url{https://arxiv.org/abs/2203.11139}

\bibitem{zhang2023lidarcamerapanopticsegmentationgeometryconsistent}
Zhang, Z., Zhang, Z., Yu, Q., Yi, R., Xie, Y., Ma, L.: Lidar-camera panoptic
  segmentation via geometry-consistent and semantic-aware alignment (2023),
  \url{https://arxiv.org/abs/2308.01686}

\bibitem{zhou2018voxelnet}
Zhou, Y., Tuzel, O.: Voxelnet: End-to-end learning for point cloud based 3d
  object detection. In: Proceedings of the IEEE conference on computer vision
  and pattern recognition. pp. 4490--4499 (2018)

\bibitem{zhou2022centerformercenterbasedtransformer3d}
Zhou, Z., Zhao, X., Wang, Y., Wang, P., Foroosh, H.: Centerformer: Center-based
  transformer for 3d object detection (2022),
  \url{https://arxiv.org/abs/2209.05588}

\end{thebibliography}
\end{document}